\documentclass[10pt, a4paper]{article}
\usepackage{amsmath}
\usepackage{multirow}
\usepackage{enumitem}
\usepackage[table]{xcolor}
\usepackage{amsfonts}
\usepackage{booktabs}
\usepackage{caption}
\usepackage{hyperref}
\usepackage[final]{lrec2026} 

\newcommand{\gloss}[2]{\href{#2}{\textsc{#1}}}

\title{How Pragmatics Shape Articulation: \\
A Computational Case Study in STEM ASL Discourse}

\name{
\begin{tabular}{c}
\textbf{Saki Imai}$^{1}$, 
\textbf{Lee Kezar}$^{2}$, 
\textbf{Laurel Aichler}$^{2}$, 
\textbf{Mert İnan}$^{1}$, \\
\textbf{Erin Walker}$^{3}$, 
\textbf{Alicia Wooten}$^{2}$, 
\textbf{Lorna Quandt}$^{2}$, 
\textbf{Malihe Alikhani}$^{1}$
\end{tabular}
}

\address{
$^{1}$Northeastern University \quad
$^{2}$Gallaudet University \quad
$^{3}$University of Pittsburgh 
}

\abstract{
Most state-of-the-art sign language models are trained on interpreter or isolated vocabulary data, which overlooks the variability that characterizes natural dialogue. However, human communication dynamically adapts to contexts and interlocutors through spatiotemporal changes and articulation style. This specifically manifests itself in educational settings, where novel vocabularies are used by teachers, and students. To address this gap, we collect a motion capture dataset of American Sign Language (ASL) STEM (Science, Technology, Engineering, and Mathematics) dialogue that enables quantitative comparison between dyadic interactive signing, solo signed lecture, and interpreted articles. Using continuous kinematic features, we disentangle dialogue-specific entrainment from individual effort reduction and show spatiotemporal changes across repeated mentions of STEM terms. On average, dialogue signs are 24.6\%-44.6\% shorter in duration than the isolated signs, and show significant reductions absent in monologue contexts. Finally, we evaluate sign embedding models on their ability to recognize STEM signs and approximate how entrained the participants become over time.
Our study bridges linguistic analysis and computational modeling to understand how pragmatics shape sign articulation and its representation in sign language technologies.
 \\ \newline  }

\begin{document}

\maketitleabstract

\section{Introduction}
Human communication is inherently adaptive \cite{clark1991grounding}, causing context-sensitive phenomena such as \textbf{entrainment}, where speakers adjust their linguistic and articulatory behaviors to one another \cite{Brennan}, and \textbf{effort reduction}, where individuals reduce articulatory effort through repetition \cite{zipf2016human}.
While work in dialogue modeling in spoken languages has progressed in recent years, there have been few efforts to model these processes in signed languages, which have important features related to adaptivity, such as sign lowering \cite{tyrone2010sign}, weak drop \cite{padden1987american}, and location undershooting \cite{mauk2003undershoot}, distinguishing them from spoken languages.
Models aiming to understand or generate naturalistic signing should accommodate these idiosyncratic, context-dependent articulatory shifts.

\begin{figure}[!ht]
    \centering
    \includegraphics[width=1.07\linewidth]{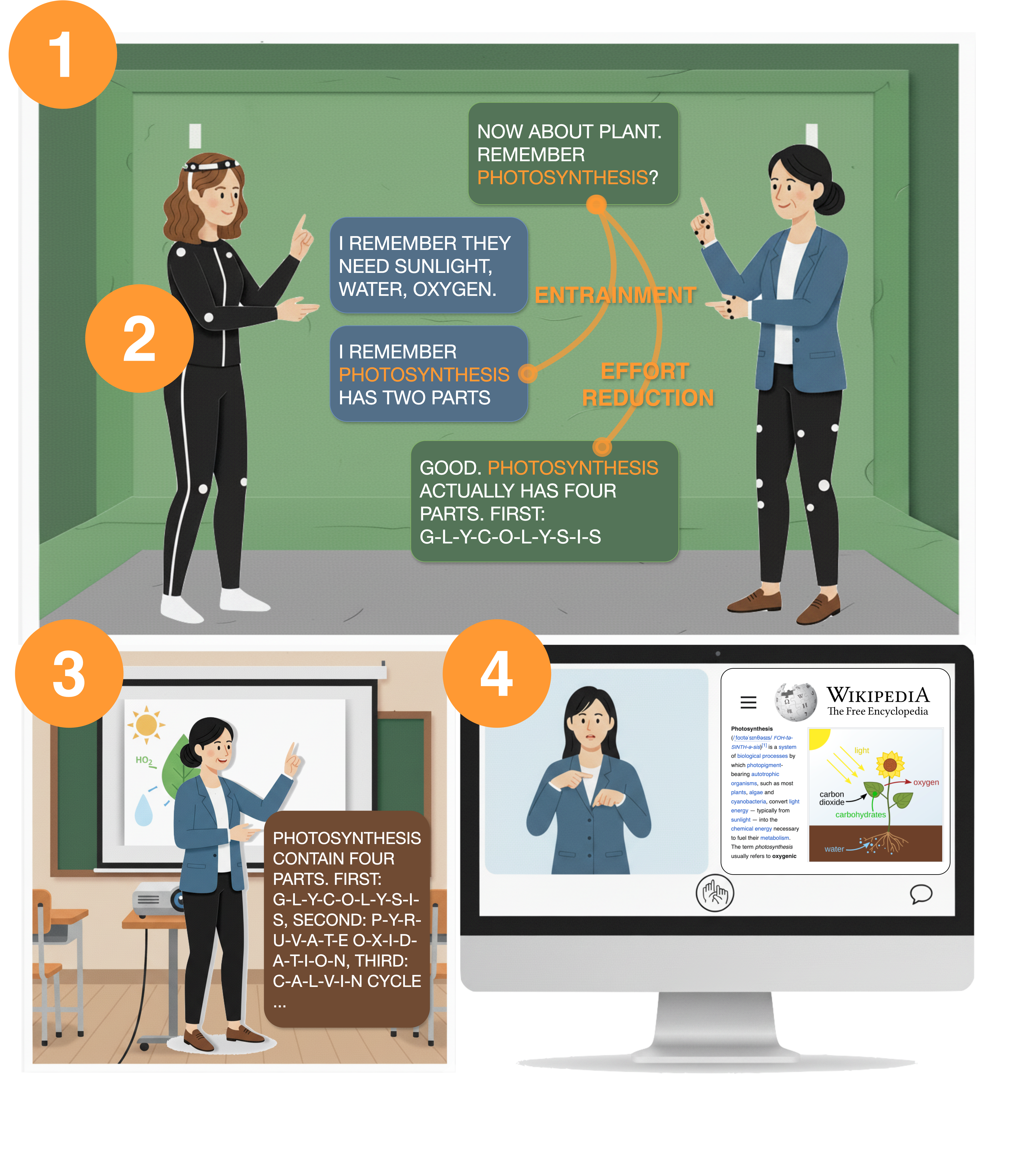}\vspace{-25pt}
    \caption{We explore the effects of pragmatics in four different signed STEM contexts: 1) instructor-student dyadic conversations, 2) isolated vocabulary, 3) a signed lecture, and 4) interpreted Wikipedia articles. We computationally analyze how signers establish conceptual pacts across these contexts.}
    \vspace{-15pt}
    \label{fig:fig1}
\end{figure}

Yet, most existing sign language models are trained on interpreter data or isolated sign videos \cite{desai-etal-2024-systemic}.
While these datasets capture clear, standardized productions, they omit the adaptive and fluid behaviors that characterize dialogue, and often reflect the influence of non-native interpreters \cite{tanzer-etal-2024-reconsidering}. 
To explore these, we test \textit{whether language models generalize beyond isolated sign data to capture interaction-driven articulatory variation.} Addressing this question requires quantifying how signers adapt their articulation in dialogue, and determining whether existing models are sensitive to these pragmatic differences.

To study this, we collect an American Sign Language (ASL) motion capture dataset that consists of two contexts: 1) an instructor-student dialogue featuring repeated biology signs used in an interactive setting, and 2) isolated productions of biology vocabulary signs, each produced individually by the student.\footnote{Data can be made available to researchers upon proper agreements.}
We supplement these with 3) publicly available recordings featuring a lecture given by the same instructor (to control for individual effort reduction), and 4) interpreter signing (representative of model training data).
This approach (Fig.~\ref{fig:fig1}) enables us to test whether dialogue exhibits additional convergence beyond per-signer efficiency and to compare natural communicative articulation with the controlled productions that dominate existing signed corpora.

We focus on biology signs because STEM signs in general represent a unique domain for studying communicative adaptation, especially in ASL. Many of these STEM concept signs lack one standardized sign form, with multiple variants circulating among scientists, interpreters, and educators \cite{lualdi2023advancing}. For instance, ASL users in some schools may sign \gloss{cell}{https://www.aslcore.org/biology/entries/?id=cell} one way, while students in another region or school may typically fingerspell \textsc{cell} letter-by-letter, or use a different sign variation, for example \gloss{cell}{http://aslsignbank.haskins.yale.edu/dictionary/gloss/4215.html} (version 2). This fluidity introduces challenges for technologies with specialized vocabulary: systems trained on particular sign forms may fail to recognize the variants used by groups of learners and educators. 

Our goal is not to generalize across all ASL use, but to introduce computational tools for quantifying sign articulation, situate articulatory findings from ASL STEM dialogue relative to prior linguistic work, and evaluate whether current sign embedding models capture these context dependent differences that arise in interactive signing. Our contributions are as follows:

\begin{enumerate}[noitemsep, nolistsep]
    \item We introduce a framework that isolates dialogue-specific entrainment from individual effort reduction in ASL, using continuous spatial, temporal, and vertical motion metrics (\S~\ref{ssec:metrics}).
    \item We find that dialogue signing exhibits reduced spatial and temporal articulation compared to isolated vocabulary productions, reflecting adaptive changes characteristic of interactive communication (\S~\ref{ssec:vocab_comparison}). 
    \item We show statistically significant spatial and temporal reduction in STEM signs across repeated mentions in dialogue, absent in monologue, indicating that such adaptation may reflect entrainment rather than individual effort reduction (\S~\ref{ssec:repeated_mention}).
    \item We demonstrate that existing sign language models fail to generalize to these articulatory differences (\S~\ref{ssec:spotting}).
\end{enumerate}
\section{Related Work}

\paragraph{Articulatory Reduction and Interactional Adaptation}
A line of research argues that spoken languages are shaped by a functional pressure toward ease of articulation and communicative efficiency \cite{zipf2016human, kanwal2017zipf, piantadosi2011word, sigurd2004word, gibson2019efficiency}. In sign languages, similar efficiency pressures shape articulatory form \cite{napoli2011some, yin-etal-2024-american, caselli2022perceptual}.
Increased signing rate, for instance, causes signs to be articulated in a lower, more compact space (a phenomenon known as sign lowering) \cite{tyrone2010sign, tyrone2012phonetic, mauk2012location}.
Moreover, casual signing tend to involve less joint usage and a shift toward distal articulators \cite{napoli2014linguistic, stamp2022capturing}.
Beyond individual efficiency, interactive contexts introduce an additional adaptation.
In dialogue, signers not only economize their own movements but may also modulate them to align more closely with their interlocutor, a process known as entrainment \cite{hoetjes2014repeated}, which recent sign language corpora are beginning to capture \cite{bono-etal-2020-utterance, bono-etal-2024-data}.
Despite these insights, characterizations of articulatory adaptation in sign languages remain limited.
Prior work has primarily relied on categorical annotation of sign which does not capture the continuous dynamics of articulatory change.
In contrast, our work employs motion capture kinematic features to disentangle individual effort reduction from dialogue specific entrainment, building on prior work with motion capture \cite{lu_pengfei, huenerfauth2010eliciting}.

\paragraph{Sign Language Modeling and Dataset Biases}
Despite recent progress in sign language recognition, most computational models remain limited in their representation of articulatory variation. Existing large-scale datasets such as PHOENIX-2014T \cite{Camgoz_2018_CVPR}, How2Sign \cite{duarte2021how2sign}, CSL-Daily \cite{zhou2021improving}, and WMTSLT \cite{muller_mathias_2022_6621480}, designed primarily for sentence-level translation tasks, often feature interpreter-produced signing performed under controlled conditions.
Consequently, this results in a strong bias toward isolated sign articulation, which diverges from the fluid and adaptive behaviors observed in dialogue. Prior work has documented instances where deaf viewers struggle to fully understand such interpretations \cite{alexander2022news}. Moreover, the dominant paradigm of sign-to-text translation implicitly treats sign language as a visual analog of spoken language rather than representations grounded in sign-specific discourse structure \cite{tanzer-etal-2024-reconsidering}. Addressing these limitations requires datasets and modeling frameworks that account for the continuous and discourse-dependent variability of natural signing. Our work contributes to this direction by introducing motion capture analyses of naturalistic ASL dialogue and by testing whether existing sign embedding models capture articulatory differences that emerge through interaction.

\section{Methods}
In this section, we describe our method of collecting and annotating the STEM dialogue data (\S\ref{ssec:data_collection}) and finding comparable monologic data (\S\ref{ssec:additional_data}). Then, we describe our analysis of the data from the perspective of motion capture kinematics (\S\ref{ssec:metrics}) and pretrained machine learning models (\S\ref{ssec:encoder_models}).

\subsection{Data Collection}
\label{ssec:data_collection}
\paragraph{Participants}
Two fluent deaf signers participated in a structured dialogue with one another.
One signer is a faculty member in Biology who regularly teaches biology classes in ASL.
The other signer is a student familiar with biology concepts from having taken classes with the faculty member previously.
Both signers are right-handed. 

\paragraph{Signed Content}
The motion capture data were collected in two conditions: 
\begin{enumerate}[noitemsep, nolistsep]
    \item \textit{Isolated STEM vocabulary articulation.} One signer (the student described above) produced a set of 77 STEM signs drawn from introductory biology content. Each sign was produced individually, without a conversational partner. This condition serves as a articulatory baseline, which allows us to estimate each sign's spatial and temporal properties before interactional adaptation occurs. The full list of STEM signs used is provided in the Appendix~\ref{app:full_STEM_list}.
    \item \textit{Biology dialogue.} The two participants engaged in an 8.52-minute spontaneous instructor-student dialogue focusing on key biology topics such as cell structure and genetics, cell cycle, and photosynthesis. While the instructor initiated inquiries to guide the interaction, the dialogue also included extended explanations, clarifications, confirmations, and belief exchanges. Across the dialogue, 17 STEM signs overlapped with the vocabulary condition, allowing for a direct comparison of articulatory variation between isolated and interactive contexts.
\end{enumerate}

\paragraph{Recording Set Up}
Both motion capture data were recorded at 120 frames per second using a Vicon system.
The setup employed 18 high-resolution cameras (8 T160 and 10 Vero) and 73 markers carefully placed on the signer’s fingers, hands, and body. Data were captured using Vicon Shogun 1.7, a tool that significantly enhanced the quality of motion capture, particularly in capturing body, hand, and finger movements with high fidelity. In addition to motion capture, we also recorded a 2D RGB video which was used for annotation.

\begin{figure*}
    \centering
    \includegraphics[width=\linewidth]{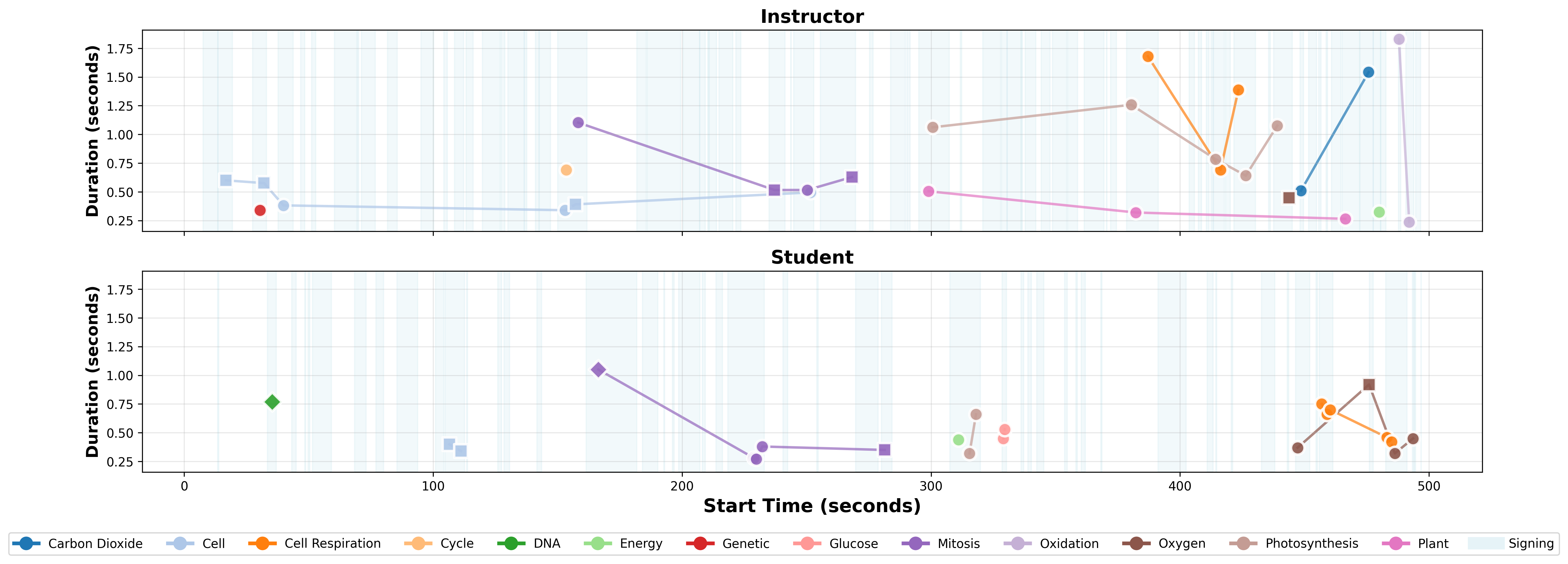}
    \caption{Sign duration patterns over time for instructor-student ASL dialogue. Stacked panels show start time (x) vs. duration (y) of individual signs for both participants, with connected lines indicating temporal progression of repeated signs within semantic groups. Each point represents an individual STEM sign instance, colored by semantic base term and shaped by articulation variation (circles, squares, diamonds for fingerspelling). Light blue background shading indicates active signing periods for each participant.}
    \label{fig:summary}
\end{figure*}

\paragraph{Annotation}
\label{ssec:annotation}
We annotated our 2D video data using ELAN \citeplanguageresource{elan2025}. ELAN uses tiers to annotate different aspects of the source language (e.g. phonology, lexical items).
Each video was annotated on two tiers: one for sentence-level translation into English and one for the STEM signs themselves. 
We followed the SLAASH ID glossing principles \citeplanguageresource{hochgesang2022slaash}, specifically focusing on accurate capture of the STEM signs to allow us to perform accurate measures of sign timing differences across the data. A gloss is a conventional written label used by researchers to reference a specific ASL sign. Two hearing, ASL-proficient researchers performed most of the annotation, consulting with the Deaf faculty member signer and a hearing fluent signer with expertise in STEM signs (Figure~\ref{fig:summary}).

\subsection{Supplementary Monologic Data}
\label{ssec:additional_data}
To contextualize the types of signing observed in STEM dialogue, we compared the collected data against \textit{monologic} ASL signing covering the same Biology topics as the dialogue: ASL STEM Wiki \citeplanguageresource{yin-etal-2024-asl} and Atomic Hands \citeplanguageresource{atomichands}.

ASL STEM Wiki provides translations of STEM articles on Wikipedia, signed by certified ASL interpreters.
From the 254 articles, we selected samples that match the topics found in the dialogue, namely, \href{https://aslgames.azurewebsites.net/wiki/?targetArticle=26310&targetName=Reproduction}{\textit{Reproduction}} and \href{https://aslgames.azurewebsites.net/wiki/?targetArticle=24544&targetName=Photosynthesis}{\textit{Photosynthesis}}.
From these two articles, we selected excerpts 8 and 12 sentences respectively\footnote{These are contiguous in the article, starting with \S5 in Reproduction and \S1 in Photosynthesis.} from the opening section to maximize lexical overlap with the dialogue. This dataset represents interpreter signing that closely resembles the dominant training data for sign language understanding models. It therefore serves as a reference point for evaluating how dialogue signing diverges from data that shape existing technologies.

With permission from the creators, we additionally analyze videos from Atomic Hands website \cite{atomichands}.
Atomic Hands provides educational resources for STEM topics in ASL, signed by deaf experts in their disciplines.
We found one video (``\href{https://atomichands.com/videos/asl-signs-cell-division-mitosis-and-meiosis/}{ASL Signs – cell division, mitosis, meiosis}'') that discusses the relevant topics in the dialogue.
The signer in this video is the same participant as the instructor in the dialogue. This dataset allows to compare dialogue signing against solo instructional signing by the same individual. This comparison helps distinguish individual effort reduction tendencies from entrainment effects that arise uniquely in dialogue.

We repeat the annotation process on these videos and include them in our experiments and analyses.

\paragraph{Extracting Keypoints}
While the motion capture recordings provided 3D joint coordinates (as described in \S~\ref{ssec:data_collection}), the additional video data consisted of 2D RGB recordings in MP4 format. To obtain a comparable set of kinematic features from these videos, we processed them using MediaPipe Holistic \cite{zhang2020mediapipe}, which extracts 543 landmarks spanning face, hands, and body. We used the default detection and tracking confidence thresholds of 0.5. We then mapped MediaPipe's landmark indices to the corresponding motion capture joint positions used in our analyses. The full list of joint correspondences between MediaPipe and the motion capture joints is provided in Appendix~\ref{app:joint_mapping}.

\subsection{Articulatory Metrics}
\label{ssec:metrics}
To quantify articulatory variability in ASL production, we adapted and extended kinematic measures commonly used in gesture studies. These metrics capture changes in spatial, temporal, and vertical movements across repeated sign articulations. We use these metrics to separate (i) \textit{individual effort reduction}, where signers economize movement through repetition regardless of interaction, from (ii) \textit{interaction-driven entrainment}, where articulatory patterns adapt in response to an interlocutor. Across all metrics, \textit{reduction} is operationalized as decreases in movement magnitude, duration, or spatial extent across repeated mentions. When such reductions occur in dialogue but are absent or attenuated in monologue by the same signer, we interpret them as evidence of entrainment rather than general articulatory efficiency. Conversely, reductions observed in both dialogue and monologue are interpreted as reflecting individual effort reduction.

\subsubsection{Spatial Reduction}
We operationalize the spatial properties of sign articulation using two measures:

\paragraph{Spatial Extent} captures the overall 3D volume occupied by a sign. For a trajectory 
$\mathbf{p}_i = (x_i, y_i, z_i)$ over frames $i=1,\dots,n$, we compute the per-dimension ranges
$\Delta x = \max_i x_i - \min_i x_i$, 
$\Delta y = \max_i y_i - \min_i y_i$, and 
$\Delta z = \max_i z_i - \min_i z_i$. 
The spatial extent is defined as the diagonal length of the corresponding bounding box:
\[
\text{SpatialExtent} = \sqrt{(\Delta x)^2 + (\Delta y)^2 + (\Delta z)^2}.
\]
This metric captures how much space an articulator occupies during signing \cite{bevacqua2006multimodal}.

\paragraph{Path Length} measures the cumulative distance traveled by the articulator across frames \cite{shaw, trujillo2020communicative}. For consecutive positions $\mathbf{p}_i, \mathbf{p}_{i+1}$:
\[
d_i = \sqrt{(x_{i+1}-x_i)^2 + (y_{i+1}-y_i)^2 + (z_{i+1}-z_i)^2},
\]
and the total path length is $\sum_{i=1}^{n-1} d_i$.

\subsubsection{Temporal Reduction}
Temporal reduction reflects efficiency in the timing of sign articulation:

\paragraph{Average Velocity} captures the mean speed of the articulator over the course of a sign. 

\paragraph{Articulation Duration} measures the total time span of a sign. Reductions in duration across repeated mentions are commonly interpreted as efficiency gain \cite{hoetjes2014repeated}. This metric does not utilize kinematic features.

\subsubsection{Sign Lowering}
Sign lowering captures the vertical displacement of the hands, often observed as signs being produced progressively lower in signing space over time or across repetitions \cite{tyrone2010sign, tyrone2012phonetic, mauk2012location}. For each sign instance, we calculate the mean vertical position of the dominant hand during articulation.

\subsection{Pre-Trained Sign Encoders and Evaluation Setup}
\label{ssec:encoder_models}
We next assess whether current state-of-the-art encoder models for ASL can generalize beyond isolated sign data to the articulatory variation found in dialogue.
Such models are commonly trained on data that is not representative of signing among deaf and hard-of-hearing people \cite{desai-etal-2024-systemic}, such as isolated sign productions and hearing signers of varying skill levels.
In contrast, our eight-minute biology dialogue presents a realistic conversation between two deaf signers, using out-of-vocabulary STEM signs while also adapting those signs through reduction, coarticulation, and entrainment.
By testing pre-trained encoders in this domain, we ask whether their embedding spaces remain stable across contextual variation and whether they can still recover lexical identity when signs deviate from canonical form.

\paragraph{Models}
We examine two pre-trained encoders with complementary input modalities and training regimes: SignCLIP and I3D.
\textbf{(1) SignCLIP} \citeplanguageresource{jiang-etal-2024-signclip} uses MediaPipe skeletal keypoints as input and encodes sign clips via a frozen 3D CNN video encoder paired with a BERT-based text encoder \cite{bert}. It is trained with a CLIP-style contrastive loss \cite{clip} to align signs and glosses using ~500k sign videos from 44 sign languages in the SpreadTheSign dictionary \cite{spreadthesign}. We use the ASL finetuned checkpoint\footnote{\url{https://drive.google.com/drive/folders/10q7FxPlicrfwZn7_FgtNqKFDiAJi6CTc}} which is further finetuned on Popsign ASL \cite{starner2023popsign}, ASL Citizen \cite{desai2023asl}, and Sem-Lex \cite{kezar2023sem}.
\textbf{(2) I3D} \citeplanguageresource{bsl_ft} is a 25M-parameter inflated 3D ConvNet (``I3D'') pretrained on Kinetics action recognition dataset \cite{kinetics_i3d} with additional transformer layers fine-tuned on 1,000 hours of BSL broadcast footage (BSL-1K; \citealt{albanie2020bsl}).
It operates on RGB video input and is trained to classify gloss labels in continuous signing\footnote{\url{https://www.robots.ox.ac.uk/~vgg/research/bslattend/data/bsl5k.pth.tar}}.

\paragraph{Creating Avatars}
The video recordings of the dialogue differ substantially from standard datasets for training sign language models. Specifically, the participants wore black motion capture suits, lowering visual contrast, and the signers were positioned at an angle rather than facing the camera directly. To bring these data closer to in-distribution signing, we used Unity \cite{unity} to render 3D avatars from the motion capture recordings. These videos preserved articulatory motion and spatial trajectories while providing consistent signer contrast and viewing angle. Example avatar renderings are shown in Appendix~\ref{app:avatar_examples}.

\subsubsection{Embedding-based Entrainment Metrics}
To explore whether pretrained encoder embeddings are sensitive to conversational adaptation, we analyze repeated STEM sign productions across time.
For each gloss $g$ with multiple occurrences per signer, we extract embedding vectors $x_1, x_2, \dots, x_T \in \mathbb{R}^d$ for each production, using mean-pooled and L2-normalized outputs from the same encoders (I3D, SignCLIP).

We define:
\begin{itemize}[noitemsep, nolistsep]
    \item $\Delta\text{cos} = \text{cos}(x_T^A, x_T^B) - \text{cos}(x_1^A, x_1^B)$: change in cross-signer similarity between first and last tokens.
    \item $s_{\text{A}\rightarrow\text{B}}$: the slope of $\text{cos}(x_i^{\text{A}}, x_1^{\text{B}})$ over $i$, indicating whether signer A becomes more similar to signer B's initial token over time.
    \item $\text{selfsim}_\text{A} = \text{cos}(x_1^{\text{A}}, x_T^{\text{A}})$: within-signer temporal stability.
\end{itemize}

We compute these metrics for all glosses with $\geq 2$ tokens per signer under both \textit{raw video} and \textit{avatar-rendered} conditions. Positive values of $\Delta\text{cos}$ or $s_{\text{A}\rightarrow\text{B}}$ are interpreted as evidence of entrainment, while negative values suggest signer-specific differentiation or self-entrainment.

\subsubsection{Sign Spotting Task}
To test these models' generalizability, we isolate STEM productions from the dialogue (manually annotated in \S\ref{ssec:annotation}) and evaluate them as search queries in a \textit{continuous sign spotting} task, where the model searches a sentence-level sequence for matching occurrences of the query.
Success on this task could lead to downstream educational technologies, such as searching a corpus of STEM content for examples of a specific sign.
We use the isolated productions (described in \S\ref{ssec:data_collection}) as queries and search the dialogic and monologic sentences using a sliding window with width = stride = $0.5$ seconds.
We additionally consider reformatting the dialogic inputs as an avatar to reduce the noise attributed to the motion capture suits.

We implement a baseline model that ranks candidate windows by cosine similarity to the query embedding.
From this ranking, we report recall at $k = \{10, 50\}$, defined as the proportion of top-$k$ windows that overlap with the ground-truth sign with intersection over union (IoU) $\geq 0.3$.
We also report mean reciprocal rank (MRR), computed as the average of $1/\text{rank}(w)$ over all windows $w$ with $\text{IoU} \geq 0.3$.
\begin{figure*}[h]
    \centering
    \includegraphics[width=0.9\linewidth]{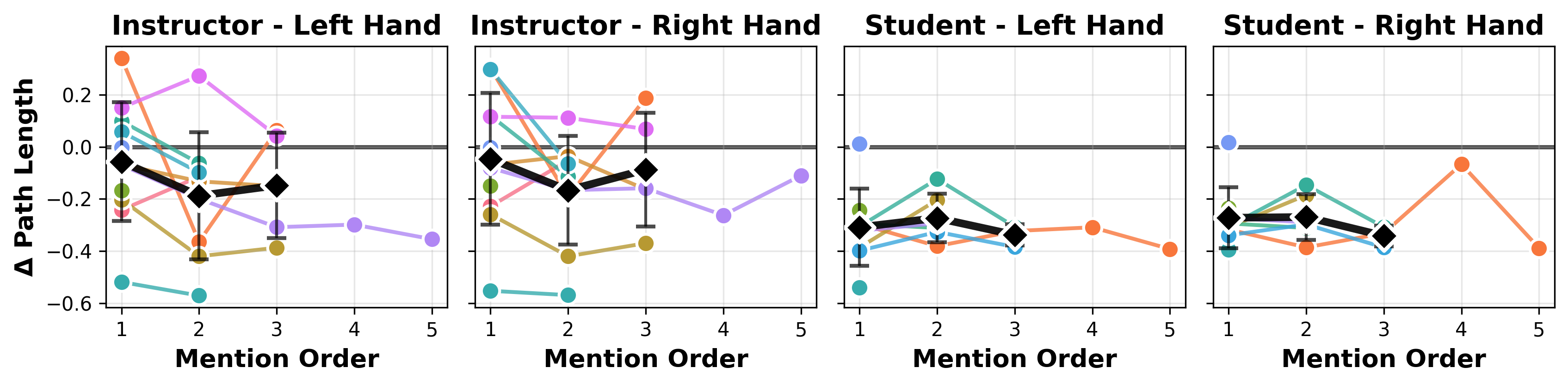}
    \caption{Path length differences between dialogue and vocabulary articulation for left and right hands as a function of mention order. Colors correspond to individual signs and the bolded black line is the mean with standard error. A value of 0 indicates no difference between dialogue and vocabulary path lengths. Negative values indicate shorter trajectories in dialogue. Overall, dialogue signing shows consistently shorter and progressively reduced movement paths relative to vocabulary signs.}
    \label{fig:path_length_delta}
\end{figure*}

\renewcommand{\arraystretch}{1.2}
\begin{table*}[ht]
\centering
\resizebox{\textwidth}{!}{
\begin{tabular}{llllllllll}
\toprule
\multirow{2}{*}{\textbf{Joint}} 
& \multicolumn{3}{c}{\textbf{Spatial Reduction}} 
& \multicolumn{3}{c}{\textbf{Path Reduction}} 
& \multicolumn{3}{c}{\textbf{Velocity Increase}} \\
\cmidrule{2-10}
& \multicolumn{1}{c}{Dialogue} & \multicolumn{2}{c}{Monologue} 
& \multicolumn{1}{c}{Dialogue} & \multicolumn{2}{c}{Monologue} 
& \multicolumn{1}{c}{Dialogue} & \multicolumn{2}{c}{Monologue} \\
\midrule
Fingers (L)  & $+0.66$*** & $-0.41$* & $(-)$  & $+0.63$***  & $-0.24$ & $(-)$ & $-0.52$** & $+0.23$ & $(-)$ \\
\rowcolor{gray!20} Fingers (R) & $-0.08$ & $-0.17$ & $(+0.54)$ & $+0.09$ & $-0.33$ & $(+0.78)$ & $+0.14$ & $+0.28$ & $(+0.45)$ \\
Hand (L)      & $+0.71$*** & $-0.29$ & $(+0.34)$ & $+0.68$*** & $-0.35$ & $(+0.34)$ & $-0.55$** & $+0.50$** & $(+0.19)$ \\
\rowcolor{gray!20} Hand (R)    & $+0.30$ & $-0.34$ & $(+0.30)$ & $+0.62$*** & $-0.39$* & $(+0.64)$ & $-0.12$ & $+0.48$** & $(+0.48)$ \\
Forearm (L)   & $+0.48$* & $-0.25$ & $(+0.73\text{*})$ & $+0.64$*** & $-0.34$ & $(+0.41)$ & $-0.36$ & $+0.41$* & $(+0.44)$ \\
\rowcolor{gray!20} Forearm (R) & $-0.04$ & $-0.38$* & $(+0.66)$ & $+0.24$ & $-0.41$* & $(+0.66)$ & $-0.05$ & $+0.51$** & $(+0.57)$ \\
Arm (L)      & $+0.44$* & $-0.23$ & $(+0.40)$ & $+0.50$* & $-0.33$ & $(+0.35)$ & $-0.48$* & $+0.45$* & $(+0.06)$ \\
\rowcolor{gray!20} Arm (R)      & $+0.35$ & $+0.08$ & $(+0.62)$ & $+0.23$ & $-0.12$ & $(+0.57)$ & $+0.04$ & $+0.15$ & $(+0.66)$ \\
\bottomrule
\end{tabular}
}
\caption{Relative Change Analysis (Spearman correlation). 
Stars denote significance (* $p<.05$, ** $p<.01$, *** $p<.001$). Values in the parentheses indicate the \textit{interpreter} condition, and non-parenthesized \textit{monologue} values correspond to the solo lecture by the same instructor in the \textit{dialogue} condition. Dialogue signing shows systematic spatial and temporal reduction, particularly in the left hand and arm, while monologue and interpreter signing exhibits weaker, inverse or inconsistent trends. The Fingers (L) values for the interpreter condition could not be computed due to Mediapipe failing to extract hand keypoints.}
\label{tab:relative_change}
\end{table*}

\section{Results}
We first compare dialogue articulations with \textit{Vocabulary} baselines to characterize how signing style differs across communicative contexts (\S~\ref{ssec:vocab_comparison}). This analysis primarily uses left and right hand motion capture data, as the hands are the primary articulators that account for the majority of articulatory effort and spatial contrast in signing. The subsequent section expands the analysis to include more joints to examine how these articulatory properties evolve across repeated mentions within the dialogue, and how that differs from a monologue context, such as solo lecture and interpreted articles (\S~\ref{ssec:repeated_mention}). Finally, we evaluate whether current sign language models can capture these lexical variations through \textit{continuous sign spotting} (\S~\ref{ssec:spotting}) experiment, and whether those embeddings are sensitive to articulatory adaptation in dialogue (\S~\ref{ssec:embedding_entrainment}).

\subsection{Comparison with Vocabulary Articulation}
\label{ssec:vocab_comparison}
To quantify how dialogue signing diverges from isolated vocabulary form and how it changes with repetition, we computed per-sign differences in kinematic properties defined in Section~\ref{ssec:metrics} between \textit{Dialogue} data and \textit{Vocabulary} conditions. In our setup, the vocabulary signs produced by the student serve as the baseline reference for both participants. Each plotted trajectory (Figures \ref{fig:path_length_delta}) shows the difference in a given metric relative to its vocabulary baseline on the y-axis, as a function of mention order on the x-axis. Thus, points below zero indicate more compact articulation than the vocabulary form, and the slope across mentions indicates how this deviation evolves with repetition. 

\paragraph{Path Length Difference}
As shown in Figure \ref{fig:path_length_delta}, most signs show negative and decreasing $\Delta$ Path length across mention order. Although there are clear inter-participant differences, these patterns appear modulated by the communicative goals of each role. The instructor shows relatively stable articulation across repetitions to ensure clarity, whereas the student who often reuses STEM terms already introduced by the instructor 
shows greater reduction from vocabulary form. The average trajectory line for each plot suggests a general downward trend, but large standard deviations show that the extent of reduction varies substantially by lexical item and interactional context. 

\paragraph{Duration Difference}
Across all STEM signs with matching vocabulary baselines, signs were on average \textbf{24.6\%} shorter for the instructor, and \textbf{44.6\%} shorter for the student, relative to their corresponding vocabulary articulation, both statistically significant reductions ($p<.01$). These results suggest that interactive signing involves not only spatial contraction but also temporal compression, which is consistent with efficiency-driven adaptation 
\cite{tyrone2010sign, hoetjes2014repeated}. 


\paragraph{Sign Lowering}
No significant differences were found in vertical hand position for either hand ($p>.2$), suggesting that while dialogue articulation exhibits clear spatial and temporal reduction, it does not consistently involve a downward shift relative to vocabulary baselines. This aligns with prior findings that sign lowering is highly context dependent and influenced by various factors such as sentence position, coarticulatory effects and prosody \cite{tyrone2012phonetic}.

\subsection{Repeated Mention Analysis}
\label{ssec:repeated_mention}
To quantify articulatory reduction across repeated mentions of a sign, we computed relative percentage change using the first occurrence of each sign as a baseline. For every sign that appeared at least twice in the corpus, we extracted joint-level motion features for all mentions and calculated the proportional reduction of each subsequent token relative to its first mention. This approach controls for signer-specific baseline differences in articulation while comparing within-sign trajectories across repetitions.

For each joint, we then tested whether the degree of reduction increased systematically with mention order using Spearman rank correlations \cite{spearman1961proof} between mention index and percent reduction. Positive correlations indicate reduction (i.e., smaller spatial extent or path length, or faster velocity) as a function of repetition. Table~\ref{tab:relative_change} reports the resulting correlations for the instructor's productions in the dialogue (\S~\ref{ssec:data_collection}) and monologue (\S~\ref{ssec:additional_data}) contexts. Comparing with the monologue condition (of the same instructor) allows us to study whether articulatory reduction across repetitions arises from individual tendencies toward effort reduction or from adaptation to an interlocutor. Results for the student signer are included in Appendix~\ref{app:student_results}, as the limited number of repeated signs in the dialogue (see Fig.~\ref{fig:summary}) reduced the statistical power of within-sign analyses.

\subsubsection{Spatial Reduction} 
\paragraph{Spatial Extent and Path Length}
Both spatial extent and path length showed systematic reduction across repeated mentions in the dialogue condition. The instructor showed significant positive correlations between mention order and reduction for several left-hand joints. This pattern may suggest a phenomenon known as \textit{weak drop} in which one hand (typically the non-dominant hand) is omitted during production of a two handed sign \cite{padden1987american}. 

In contrast, the monologue condition showed little evidence of spatial reduction and, in some cases, negative correlations.
This likely reflects the communication goals of lecture-style signing, where clarity and precision take precedence over articulatory economy.

\begin{figure}[t!]
    \centering
    \includegraphics[width=0.9\linewidth]{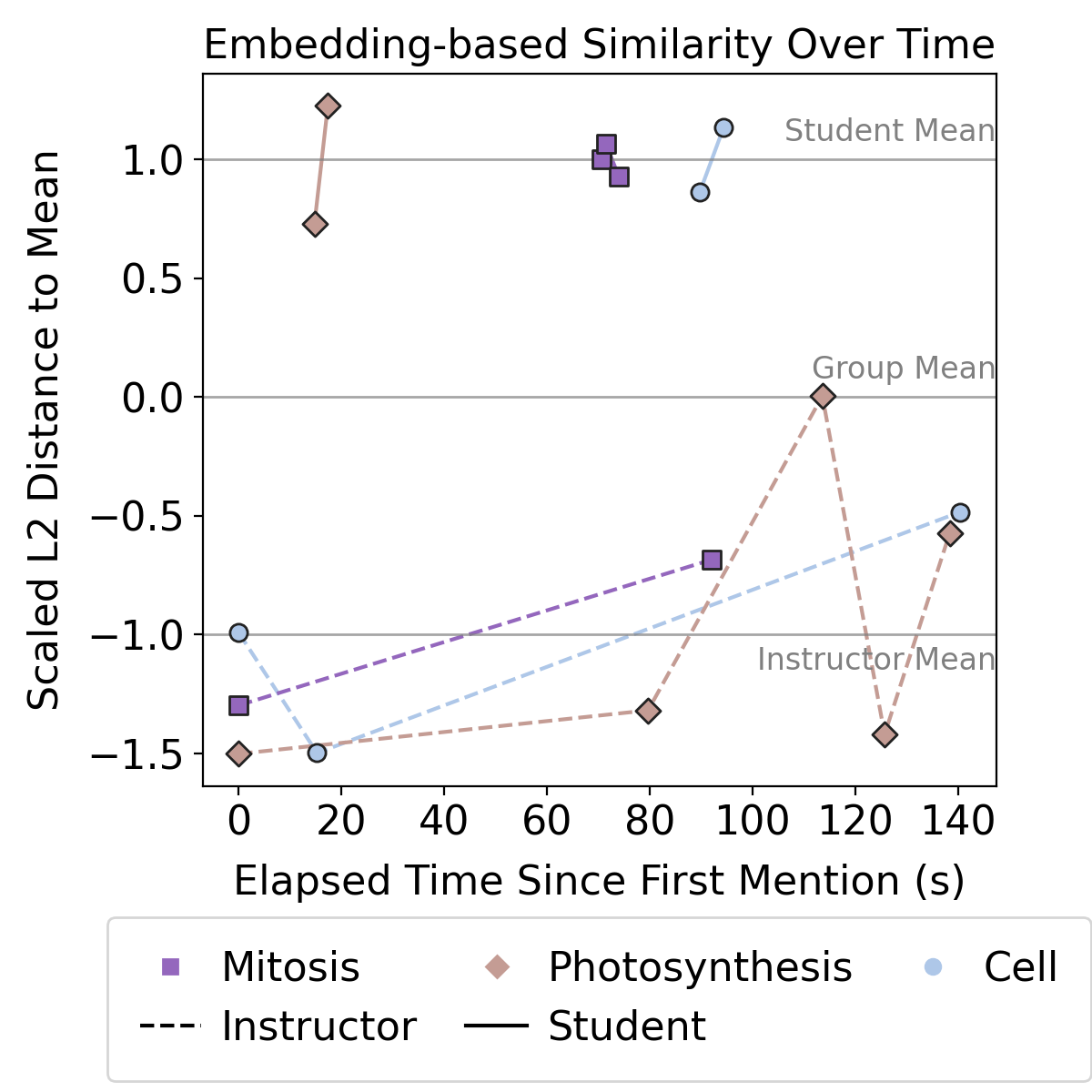}
    \caption{Sign productions plotted with respect to their embeddings' L2 distance to the mean of each signer's average and the group's average.}
    \label{fig:model_entrainment}
\end{figure}

\subsubsection{Temporal Reduction}
\paragraph{Average Velocity}
Temporal metrics revealed similar patterns. In the dialogue condition, several left-side joints exhibited significant negative correlations between motion order and velocity increase. In contrast, positive correlations were found for some right-side joints in the monologue, which may reflect emphasis through larger and more deliberate motions to maintain clarity. 

\paragraph{Articulation Duration}
In the dialogue condition, the instructor's sign durations decreased significantly with repetition ($r=0.279, p <.001$), indicating that signs become approximately $27.9\%$ shorter on average over repeated mentions.
The student data followed a similar trend ($r=0.295, p<.001$).
These reductions across interlocutors suggest a  temporal entrainment effect, where both signers adapt their articulatory timing patterns over the course of interaction \cite{brennan1996lexical, garrod1987saying}.
In the monologue, by contrast, no significant duration change was found ($r=0.026, p=0.895)$.
The absence of reduction when signing alone further suggests that the observed compression in dialogue arises from interactional adaptation rather than articulatory efficiency.


\renewcommand{\arraystretch}{1.2}
\begin{table}[t!]
    \centering
    \begin{tabular}{clrrr}
        \toprule
        \textbf{Input} & \multicolumn{1}{c}{\textbf{Model}} & \textbf{MRR} & \textbf{R@10} & \textbf{R@50} \\
        \midrule
        \rowcolor{gray!20} 
         & I3D        & $0.009$ & $0.043$ & $0.087$ \\
        \rowcolor{gray!20}
        \multirow{-2}{*}{$D_{\text{inst.}}$} & SignCLIP                      & $0.004$ & $0.000$ & $0.000$ \\
        
          & I3D        & $0.003$ & $0.000$ & $0.000$ \\
        \multirow{-2}{*}{$D_{\text{inst.}}^{\Diamond}$} & SignCLIP                      & $0.003$ & $0.000$ & $0.000$ \\
        
        \rowcolor{gray!20}  & I3D        & $0.013$ & $0.048$ & $0.095$ \\
        \rowcolor{gray!20} \multirow{-2}{*}{$D_\text{stud.}$} & SignCLIP                      & $0.003$ & $0.000$ & $0.000$ \\
        
          & I3D        & $0.011$ & $0.048$ & $0.095$ \\
        \multirow{-2}{*}{$D_{\text{stud.}}^{\Diamond}$} & SignCLIP                      & $0.013$ & $0.048$ & $0.095$ \\
        
        \rowcolor{gray!20}  & I3D        & $0.026$ & $0.086$ & $0.229$ \\
        \rowcolor{gray!20} \multirow{-2}{*}{$M_{\text{inst.}}$} & SignCLIP                      & $0.021$ & $0.029$ & $0.171$ \\
         & I3D        & $0.005$ & $0.000$ & $0.071$ \\
        \multirow{-2}{*}{$M_\text{wiki}$} & SignCLIP                      & $0.010$ & $0.000$ & $0.214$ \\
        \bottomrule
    \end{tabular}
    \caption{Retrieval performance (mean reciprocal rank, recall@$k$) by input context and encoder. Key: $D$=dialogue, $M$=monologue, $\Diamond$=avatar rendering.}
    \label{tab:spotting}
\end{table}

\subsection{Embedding-Based Entrainment}
\label{ssec:embedding_entrainment}
Having examined the effect of pragmatic context using kinematic measures, we now evaluate the learned representations from pre-trained sign language models. 

We focus on the SignCLIP encoder, excluding I3D due to near-zero variation across time ($\Delta\cos < 0.0002$).
This is likely a result of the model's training on frontal videos, which mismatches the side-facing camera perspective in our dataset.
By contrast, SignCLIP operates on normalized pose inputs.
Three signs met the minimum requirement of $n >1$ productions for each signer: \gloss{mitosis}{https://www.aslcore.org/biology/entries/?id=mitosis}, \gloss{photosynthesis}{https://www.aslcore.org/biology/entries/?id=photosynthesis}, and \gloss{cell}{https://www.aslcore.org/biology/entries/?id=cell}.

All signs exhibit a clear increase in cross-signer similarity from the first to the last repetition, with $\Delta\cos > 0$ for all three (\gloss{cell}{https://www.aslcore.org/biology/entries/?id=cell}: $+0.0070$, \gloss{mitosis}{https://www.aslcore.org/biology/entries/?id=mitosis}: $+0.0422$, \gloss{photosynthesis}{https://www.aslcore.org/biology/entries/?id=photosynthesis}: $+0.0785$), providing limited evidence of entrainment.

The student's self-similarity is consistently higher than the instructor's across all signs with average scores of $\text{selfsim}_\text{stud.} = 0.963$ and $\text{selfsim}_\text{inst.} = 0.942$ across the three signs.
This indicates that the student’s productions remained more internally stable across time.

Slope-based analyses show that the student tends to diverge slightly from the instructor $s_{\text{stud.}\rightarrow\text{inst.}} < 0$, with slopes of $-0.0329$ for \gloss{photosynthesis}{https://www.aslcore.org/biology/entries/?id=photosynthesis}, $-0.0022$ for \gloss{cell}{https://www.aslcore.org/biology/entries/?id=cell}, and $+0.0039$ for \gloss{mitosis}{https://www.aslcore.org/biology/entries/?id=mitosis}.
Instructor-to-student slopes are consistently positive, with slopes of $5e^{-4}, 9e^{-4}, 1e^{-4}$ respectively.

Figure~\ref{fig:model_entrainment} visualizes each embedding $x_t$’s alignment relative to the mean embeddings of the student $\mu_{\text{stud.}}$ and instructor $\mu_{\text{inst.}}$.
To quantify this alignment along a single axis, we compute a similarity score by projecting $x_t$ onto the line that connects the signer means, and scale to $[-1,1]$:
\[
\text{sim}(x_t) = \frac{2(x_t - \mu_{\text{inst.}})^\top (\mu_{\text{stud.}} - \mu_{\text{inst.}})}{\|\mu_{\text{stud.}} - \mu_{\text{inst.}}\|^2}-1
\]
The student's embeddings generally move \textit{away} from the instructor’s mean (by a small amount) while the instructor’s move \textit{toward} the student’s.
Taken together, these patterns suggest that the instructor adapted more to the student’s signing than vice versa, consistent with instructor-side entrainment rather than student convergence.
However, these findings are limited by the models' performance on recognition tasks, like sign spotting.

\subsection{Sign Spotting}
\label{ssec:spotting}
We evaluate sign spotting performance using pretrained encoders across the four contexts.
Table~\ref{tab:spotting} reports MRR and recall at $k=\{10,50\}$.

Performance varies widely across signer and context.
The highest scores are observed in the Instructor-monologue condition, with I3D achieving the top MRR of $0.026$ and SignCLIP close behind at $0.021$.
This aligns with expectations, as the monologue setting most closely resembles the direct-facing data seen during model pretraining.
In contrast, the dialogue settings consistently lead to lower performance.

Searching over the student's signing yields slightly better results than the instructor's, with I3D scoring $0.013$ MRR in the former compared to $0.009$ in the latter.
This asymmetry supports the hypothesis that query and search embeddings drawn from the same signer are more similar in latent space than embeddings drawn across signers.
SignCLIP, however, performs poorly in both dialogue contexts, failing to retrieve any correct items within the top-50 ranks.
When the student's signing is rendered with an avatar, the SignCLIP model gains $0.01$ MRR, but in all other cases the avatar reduced performance.

Overall, the findings confirm that signer identity and pragmatic context heavily influence retrieval outcomes.
No encoder achieves strong performance across all conditions, suggesting that current sign-receptive models do not generalize well to lexical innovations, such as in STEM contexts.
Furthermore, high performance in monologue but low recall in dialogue highlights the need for models that are robust to pragmatic variation and signer-specific features.
\section{Discussion}
This work provides the first quantitative evidence that pragmatic adaptation in sign language follows measurable articulatory principles comparable to spoken dialogue
\cite{zipf2016human, kanwal2017zipf, piantadosi2011word}. Signs produced in dialogue were both spatially and temporally reduced relative to isolated sign articulations, with reductions intensifying across repeated mentions. These findings mirror the functional pressures toward articulatory economy observed in spoken language \cite{zipf2016human, lindblom1990explaining}. 

Significant reductions in left hand and arm movements indicate selective economization of non-dominant articulators, consistent with the weak drop phenomenon \cite{padden1987american}. This suggests that pragmatic adaptation in sign language is not a uniform compression but a targeted modulation that preserves communicative clarity while minimizing redundant effort. Additionally, different patterns in the instructor and the student's signing suggest that communicative goals and power dynamics may exert pressure upon signed productions. We observed that the instructor was more likely to sign slowly and match the student's signing. The instructor may prioritize clarity and match the student's articulations, while the student aims for conciseness and demonstrates a lesser tendency to align with the instructor's signing.

From a computational perspective, our findings highlight the gap between linguistic adaptation and machine representation. Although current sign embedding models such as SignCLIP and I3D achieve high accuracy on interpreter and isolated sign benchmarks, their performance degrades in interactive settings that feature pragmatic variation. This suggests that these models largely capture lexical and visual regularities but fail to encode the gradient articulatory variability characteristic of real-world signing.

Our results highlight a broader limitation in how current sign language technologies conceptualize variation: models are trained to recognize the sign produced, but not how or why it varies. By incorporating motion capture kinematics into dialogue analysis, this study emphasizes the importance of modeling approaches that view signing as an adaptive and interactional system, rather than a sequence of static lexical items.

\section{Conclusion}
This study presented an empirical analysis of how dialogic ASL signing differs from isolated vocabulary and monologic articulations through quantitative analyses of motion capture recording in STEM discourse. Using continuous spatial, temporal, and vertical motion metrics, we found that dialogic signing is characterized by articulatory reduction, which is absent in solo lecture contexts. Our analyses further revealed that current embedding models struggle to generalize to such pragmatic variation due to the limitations of training paradigms that rely predominantly on interpreter or isolated vocabulary data.

By modeling how STEM signs are adapted in interactive ASL dialogue, our findings highlight the importance of developing educational technologies and sign language models that are robust to pragmatic variation. Future work should extend these analyses to larger datasets, diverse signers, and additional discourse contexts to better understand how communicative pressures drive articulatory adaptation across modalities.
\section*{Limitations}
\paragraph{Data Limitations}
While our dataset provides a unique and naturalistic ASL dialogue grounded in STEM education, it remains limited in scope.
The number of participants is small, consisting of a single instructor–student pair, and the duration of the recorded dialogue (8.52 minutes) restricts the range of lexical and interactional variation that can be analyzed.
For example, a different pair of interlocutors are likely to use different regional variations for STEM concepts and may show different levels of entrainment and effort reduction.
Consequently, our findings should be interpreted as a case study rather than generalization.

The video modality of the dialogue features participants wearing black body suits, which creates visual noise for both human annotators and machine learning models.
We attempted to remedy this effect with an avatar-based rendering of the motion capture data; nonetheless, both formats are considerably out-of-distribution for the pretrained models, as evidenced by our findings.

\paragraph{Annotation Challenges}
The authors who annotated the ASL data are hearing signers.
Additionally, the analyses were conducted by hearing signers and non-signers.
This lack of annotation and analysis directly from native ASL users may introduce bias into the analyses and interpretations, potentially leaving out certain valid explanations of the observed phenomena or undermining the quality of the annotations.
Proofing was conducted across both signers' annotations; however, judgments about translation correctness and sign start/stop frame are inherently subjective.

One challenge of annotating ASL and other sign languages is the three-dimensional nature of the languages and the lack of a widely accepted writing or annotation system.
Because of this gap, language researchers have historically relied on glossing ASL with words borrowed from English.
This presents several problems: 1) glosses chosen are often not consistent, leading to confusion about which sign is meant, 2) the English meaning of the word chosen often does not line up with the meaning of the ASL sign which it is supposed to represent, which can introduce skews in understanding when communicating about signs, 3) the written gloss does not show the form of the sign, leading to a lack of access for researchers who do not already sign.
The SLAASH ID Glossing Principles \cite{hochgesang2022slaash} address some of these longstanding issues. 
They emphasize the importance of bridging the distance between actual sign production and the data presented to researchers by linking annotations directly to video and always providing a video link for signs that are glossed.

\section*{Ethical Considerations}
Given that ASL is a low-resource language and the language closely associated with a minority community in the United States (and to some extent, other languages), we note the importance of having cultural and linguistic knowledge that derives from lived experience, and not only academic study. Multiple authors of this paper use ASL in their daily or workday lives and have concrete connections within Deaf communities of the United States. One author/project leader is a Deaf fluent signer, as was the "instructor" in the motion capture dataset. We recognize the inherent difficulties and weaknesses of carrying out research on minority languages with sizable research teams whose relationships to the languages vary widely, and hope that by sharing our positionality, we may be better able to ``join the conversation" regarding signed language research and emerging technologies \cite{desai-etal-2024-systemic}.

The influence of pragmatics is complex, and neither the data introduced in this work nor the analyses we conducted are fully representative of this complexity.
This limitation is especially important in educational contexts, where a diverse range of learning preferences, cultural backgrounds, and ways of languaging influence the surface form of signing.
The work presented here represents an early attempt to computationally characterize specific aspects of sign language pragmatics and inform further research on this topic.
\section*{Acknowledgement}
This research was supported in part by the U.S. National Science Foundation under Awards No. 2418662, 2418663 and 2418664. We thank Deanna Dunlop, Jason Lamberton, Thalia Guettler, and Joseph Palagano for their assistance with data collection and annotation.

\nocite{*}
\section*{Bibliographical References}\label{sec:reference}

\bibliographystyle{lrec2026-natbib}
\bibliography{lrec2026-example}

\begin{thebibliography}{6}
\expandafter\ifx\csname natexlab\endcsname\relax\def\natexlab#1{#1}\fi

\bibitem[{Hochgesang(2022)}]{hochgesang2022slaash}
Hochgesang, J. 2022.
\newblock \href {https://doi.org/https://doi.org/10.6084/m9.figshare.12003732.v4} {\emph{SLAASh ID glossing principles, ASL Signbank and annotation conventions}}.
\newblock PID \href{https://doi.org/10.6084/m9.figshare.12003732.v4}{https://doi.org/10.6084/m9.figshare.12003732.v4}.

\bibitem[{Jiang et~al.(2024)Jiang, Sant, Moryossef, M{\"u}ller, Sennrich, and Ebling}]{jiang-etal-2024-signclip}
Jiang, Zifan and Sant, Gerard and Moryossef, Amit and M{\"u}ller, Mathias and Sennrich, Rico and Ebling, Sarah. 2024.
\newblock \href {https://doi.org/10.18653/v1/2024.emnlp-main.518} {\emph{{S}ign{CLIP}: Connecting Text and Sign Language by Contrastive Learning}}.
\newblock Association for Computational Linguistics.
\newblock PID \href{https://aclanthology.org/2024.emnlp-main.518/}{https://aclanthology.org/2024.emnlp-main.518/}.

\bibitem[{{Max Planck Institute for Psycholinguistics, The Language Archive}(2025)}]{elan2025}
{Max Planck Institute for Psycholinguistics, The Language Archive}. 2025.
\newblock \emph{ELAN (Version 7.0) [Computer software]}.
\newblock Max Planck Institute for Psycholinguistics, The Language Archive.
\newblock PID \href{https://archive.mpi.nl/tla/elan}{https://archive.mpi.nl/tla/elan}.

\bibitem[{Varol et~al.(2021)Varol, Momeni, Albanie, Afouras, and Zisserman}]{bsl_ft}
Varol, G{\"u}l and Momeni, Liliane and Albanie, Samuel and Afouras, Triantafyllos and Zisserman, Andrew. 2021.
\newblock \emph{Read and Attend: Temporal Localisation in Sign Language Videos}.
\newblock PID \href{https://arxiv.org/abs/2103.16481}{https://arxiv.org/abs/2103.16481}.

\bibitem[{Wooten and Spieker(2022)}]{atomichands}
Wooten, Alicia and Spieker, Barbara. 2022.
\newblock \emph{Home}.
\newblock PID \href{https://atomichands.com/}{https://atomichands.com/}.

\bibitem[{Yin et~al.(2024)Yin, Singh, Minakov, Milan, Daum{\'e}~Iii, Zhang, Lu, and Bragg}]{yin-etal-2024-asl}
Yin, Kayo and Singh, Chinmay and Minakov, Fyodor O and Milan, Vanessa and Daum{\'e} Iii, Hal and Zhang, Cyril and Lu, Alex Xijie and Bragg, Danielle. 2024.
\newblock \href {https://doi.org/10.18653/v1/2024.emnlp-main.801} {\emph{{ASL} {STEM} {W}iki: Dataset and Benchmark for Interpreting {STEM} Articles}}.
\newblock Association for Computational Linguistics.
\newblock PID \href{https://aclanthology.org/2024.emnlp-main.801/}{https://aclanthology.org/2024.emnlp-main.801/}.

\end{thebibliography}


\begin{thebibliography}{59}
\expandafter\ifx\csname natexlab\endcsname\relax\def\natexlab#1{#1}\fi

\bibitem[{Albanie et~al.(2020)Albanie, Varol, Momeni, Afouras, Chung, Fox, and Zisserman}]{albanie2020bsl}
Samuel Albanie, G{\"u}l Varol, Liliane Momeni, Triantafyllos Afouras, Joon~Son Chung, Neil Fox, and Andrew Zisserman. 2020.
\newblock Bsl-1k: Scaling up co-articulated sign language recognition using mouthing cues.
\newblock In \emph{European conference on computer vision}, pages 35--53. Springer.

\bibitem[{Alexander and Rijckaert(2022)}]{alexander2022news}
Dhoest Alexander and Jorn Rijckaert. 2022.
\newblock News ‘with’or ‘in’sign language? case study on the comprehensibility of sign language in news broadcasts.
\newblock \emph{Perspectives}, 30(4):627--642.

\bibitem[{Bevacqua et~al.(2006)Bevacqua, Raouzaiou, Peters, Caridakis, Karpouzis, Pelachaud, and Mancini}]{bevacqua2006multimodal}
Elisabetta Bevacqua, Amaryllis Raouzaiou, Christopher Peters, George Caridakis, Kostas Karpouzis, Catherine Pelachaud, and Maurizio Mancini. 2006.
\newblock Multimodal sensing, interpretation and copying of movements by a virtual agent.
\newblock In \emph{International Tutorial and Research Workshop on Perception and Interactive Technologies for Speech-Based Systems}, pages 164--174. Springer.

\bibitem[{Bono et~al.(2024)Bono, Okada, Skobov, and Adam}]{bono-etal-2024-data}
Mayumi Bono, Tomohiro Okada, Victor Skobov, and Robert Adam. 2024.
\newblock \href {https://aclanthology.org/2024.signlang-1.3/} {Data integration, annotation, and transcription methods for sign language dialogue with latency in videoconferencing}.
\newblock In \emph{Proceedings of the LREC-COLING 2024 11th Workshop on the Representation and Processing of Sign Languages: Evaluation of Sign Language Resources}, pages 26--35, Torino, Italia. ELRA and ICCL.

\bibitem[{Bono et~al.(2020)Bono, Sakaida, Okada, and Miyao}]{bono-etal-2020-utterance}
Mayumi Bono, Rui Sakaida, Tomohiro Okada, and Yusuke Miyao. 2020.
\newblock \href {https://aclanthology.org/2020.signlang-1.3/} {Utterance-unit annotation for the {JSL} dialogue corpus: Toward a multimodal approach to corpus linguistics}.
\newblock In \emph{Proceedings of the LREC2020 9th Workshop on the Representation and Processing of Sign Languages: Sign Language Resources in the Service of the Language Community, Technological Challenges and Application Perspectives}, pages 13--20, Marseille, France. European Language Resources Association (ELRA).

\bibitem[{Brennan and Clark(1996)}]{Brennan}
Susan Brennan and Herbert Clark. 1996.
\newblock \href {https://doi.org/10.1037/0278-7393.22.6.1482} {Conceptual pacts and lexical choice in conversation}.
\newblock \emph{Journal of Experimental Psychology: Learning, Memory, and Cognition}, 22:1482--1493.

\bibitem[{Brennan(1996)}]{brennan1996lexical}
Susan~E Brennan. 1996.
\newblock Lexical entrainment in spontaneous dialog.
\newblock \emph{Proceedings of ISSD}, 96:41--44.

\bibitem[{Camgoz et~al.(2018)Camgoz, Hadfield, Koller, Ney, and Bowden}]{Camgoz_2018_CVPR}
Necati~Cihan Camgoz, Simon Hadfield, Oscar Koller, Hermann Ney, and Richard Bowden. 2018.
\newblock Neural sign language translation.
\newblock In \emph{Proceedings of the IEEE Conference on Computer Vision and Pattern Recognition (CVPR)}.

\bibitem[{Carreira and Zisserman(2017)}]{kinetics_i3d}
João Carreira and Andrew Zisserman. 2017.
\newblock \href {https://doi.org/10.1109/CVPR.2017.502} {Quo vadis, action recognition? a new model and the kinetics dataset}.
\newblock In \emph{2017 IEEE Conference on Computer Vision and Pattern Recognition (CVPR)}, pages 4724--4733.

\bibitem[{Caselli et~al.(2022)Caselli, Occhino, Artacho, Savakis, and Dye}]{caselli2022perceptual}
Naomi Caselli, Corrine Occhino, Bruno Artacho, Andreas Savakis, and Matthew Dye. 2022.
\newblock Perceptual optimization of language: evidence from american sign language.
\newblock \emph{Cognition}, 224:105040.

\bibitem[{Clark and Brennan(1991)}]{clark1991grounding}
Herbert~H Clark and Susan~E Brennan. 1991.
\newblock Grounding in communication.

\bibitem[{Desai et~al.(2023)Desai, Berger, Minakov, Milano, Singh, Pumphrey, Ladner, Daum{\'e}~III, Lu, Caselli et~al.}]{desai2023asl}
Aashaka Desai, Lauren Berger, Fyodor Minakov, Nessa Milano, Chinmay Singh, Kriston Pumphrey, Richard Ladner, Hal Daum{\'e}~III, Alex~X Lu, Naomi Caselli, et~al. 2023.
\newblock Asl citizen: a community-sourced dataset for advancing isolated sign language recognition.
\newblock \emph{Advances in Neural Information Processing Systems}, 36:76893--76907.

\bibitem[{Desai et~al.(2024)Desai, De~Meulder, Hochgesang, Kocab, and Lu}]{desai-etal-2024-systemic}
Aashaka Desai, Maartje De~Meulder, Julie~A. Hochgesang, Annemarie Kocab, and Alex~X. Lu. 2024.
\newblock \href {https://aclanthology.org/2024.signlang-1.6/} {Systemic biases in sign language {AI} research: A deaf-led call to reevaluate research agendas}.
\newblock In \emph{Proceedings of the LREC-COLING 2024 11th Workshop on the Representation and Processing of Sign Languages: Evaluation of Sign Language Resources}, pages 54--65, Torino, Italia. ELRA and ICCL.

\bibitem[{Devlin et~al.(2019)Devlin, Chang, Lee, and Toutanova}]{bert}
Jacob Devlin, Ming-Wei Chang, Kenton Lee, and Kristina Toutanova. 2019.
\newblock \href {https://doi.org/10.18653/v1/N19-1423} {{BERT}: Pre-training of deep bidirectional transformers for language understanding}.
\newblock In \emph{Proceedings of the 2019 Conference of the North {A}merican Chapter of the Association for Computational Linguistics: Human Language Technologies, Volume 1 (Long and Short Papers)}, pages 4171--4186, Minneapolis, Minnesota. Association for Computational Linguistics.

\bibitem[{Duarte et~al.(2021)Duarte, Palaskar, Ventura, Ghadiyaram, DeHaan, Metze, Torres, and Giro-i Nieto}]{duarte2021how2sign}
Amanda Duarte, Shruti Palaskar, Lucas Ventura, Deepti Ghadiyaram, Kenneth DeHaan, Florian Metze, Jordi Torres, and Xavier Giro-i Nieto. 2021.
\newblock How2sign: a large-scale multimodal dataset for continuous american sign language.
\newblock In \emph{Proceedings of the IEEE/CVF conference on computer vision and pattern recognition}, pages 2735--2744.

\bibitem[{Garrod and Anderson(1987)}]{garrod1987saying}
Simon Garrod and Anthony Anderson. 1987.
\newblock Saying what you mean in dialogue: A study in conceptual and semantic co-ordination.
\newblock \emph{Cognition}, 27(2):181--218.

\bibitem[{Gibson et~al.(2019)Gibson, Futrell, Piantadosi, Dautriche, Mahowald, Bergen, and Levy}]{gibson2019efficiency}
Edward Gibson, Richard Futrell, Steven~P Piantadosi, Isabelle Dautriche, Kyle Mahowald, Leon Bergen, and Roger Levy. 2019.
\newblock How efficiency shapes human language.
\newblock \emph{Trends in cognitive sciences}, 23(5):389--407.

\bibitem[{Hilzensauer and Krammer(2015)}]{spreadthesign}
M.~Hilzensauer and K.~Krammer. 2015.
\newblock A multilingual dictionary for sign languages: "spreadthesign".
\newblock In \emph{ICERI2015 Proceedings}, 8th International Conference of Education, Research and Innovation, pages 7826--7834. IATED.

\bibitem[{Hoetjes et~al.(2014)Hoetjes, Krahmer, and Swerts}]{hoetjes2014repeated}
Marieke Hoetjes, Emiel Krahmer, and Marc Swerts. 2014.
\newblock Do repeated references result in sign reduction?
\newblock \emph{Sign Language \& Linguistics}, 17(1):56--81.

\bibitem[{Huenerfauth and Lu(2010)}]{huenerfauth2010eliciting}
Matt Huenerfauth and Pengfei Lu. 2010.
\newblock Eliciting spatial reference for a motion-capture corpus of american sign language discourse.
\newblock In \emph{sign-lang@ LREC 2010}, pages 121--124. European Language Resources Association (ELRA).

\bibitem[{Johnson and Liddell(2010)}]{johnson2010toward}
Robert~E Johnson and Scott~K Liddell. 2010.
\newblock Toward a phonetic representation of signs: Sequentiality and contrast.
\newblock \emph{Sign Language Studies}, 11(2):241--274.

\bibitem[{Johnson and Liddell(2011{\natexlab{a}})}]{johnson2011segmental}
Robert~E Johnson and Scott~K Liddell. 2011{\natexlab{a}}.
\newblock A segmental framework for representing signs phonetically.
\newblock \emph{Sign Language Studies}, 11(3):408--463.

\bibitem[{Johnson and Liddell(2011{\natexlab{b}})}]{johnson2011toward}
Robert~E Johnson and Scott~K Liddell. 2011{\natexlab{b}}.
\newblock Toward a phonetic representation of hand configuration: The fingers.
\newblock \emph{Sign Language Studies}, 12(1):5--45.

\bibitem[{Johnson and Liddell(2012)}]{johnson2012toward}
Robert~E Johnson and Scott~K Liddell. 2012.
\newblock Toward a phonetic representation of hand configuration: The thumb.
\newblock \emph{Sign Language Studies}, 12(2):316--333.

\bibitem[{Johnson and Liddell(2021)}]{johnson2021toward}
Robert~E Johnson and Scott~K Liddell. 2021.
\newblock Toward a phonetic description of hand placement on bearings.
\newblock \emph{Sign Language Studies}, 22(1):131--180.

\bibitem[{Kanwal et~al.(2017)Kanwal, Smith, Culbertson, and Kirby}]{kanwal2017zipf}
Jasmeen Kanwal, Kenny Smith, Jennifer Culbertson, and Simon Kirby. 2017.
\newblock Zipf’s law of abbreviation and the principle of least effort: Language users optimise a miniature lexicon for efficient communication.
\newblock \emph{Cognition}, 165:45--52.

\bibitem[{Kezar et~al.(2023)Kezar, Thomason, Caselli, Sehyr, and Pontecorvo}]{kezar2023sem}
Lee Kezar, Jesse Thomason, Naomi Caselli, Zed Sehyr, and Elana Pontecorvo. 2023.
\newblock The sem-lex benchmark: Modeling asl signs and their phonemes.
\newblock In \emph{Proceedings of the 25th International ACM SIGACCESS Conference on Computers and Accessibility}, pages 1--10.

\bibitem[{Krauss and Weinheimer(1964)}]{krauss1964changes}
Robert~M Krauss and Sidney Weinheimer. 1964.
\newblock Changes in reference phrases as a function of frequency of usage in social interaction: A preliminary study.
\newblock \emph{Psychonomic Science}, 1:113--114.

\bibitem[{Liddell(1984)}]{liddell1984think}
Scott~K Liddell. 1984.
\newblock Think and believe: sequentiality in american sign language.
\newblock \emph{Language}, pages 372--399.

\bibitem[{Liddell and Johnson(1986)}]{liddell1986american}
Scott~K Liddell and Robert~E Johnson. 1986.
\newblock American sign language compound formation processes, lexicalization, and phonological remnants.
\newblock \emph{Natural Language \& Linguistic Theory}, 4(4):445--513.

\bibitem[{Liddell and Johnson(1989)}]{liddell1989american}
Scott~K Liddell and Robert~E Johnson. 1989.
\newblock American sign language: The phonological base.
\newblock \emph{Sign language studies}, 64(1):195--277.

\bibitem[{Liddell and Johnson(2019)}]{liddell2019sign}
Scott~K Liddell and Robert~E Johnson. 2019.
\newblock Sign language articulators on phonetic bearings.
\newblock \emph{Sign Language Studies}, 20(1):132--172.

\bibitem[{Lindblom(1990)}]{lindblom1990explaining}
Bj{\"o}rn Lindblom. 1990.
\newblock Explaining phonetic variation: A sketch of the h\&h theory.
\newblock In \emph{Speech production and speech modelling}, pages 403--439. Springer.

\bibitem[{Lu and Huenerfauth(2012)}]{lu_pengfei}
Pengfei Lu and Matt Huenerfauth. 2012.
\newblock \href {https://www.sign-lang.uni-hamburg.de/lrec/pub/12005.pdf} {{CUNY} {American} {Sign} {Language} motion-capture corpus: First release}.
\newblock In \emph{Proceedings of the {LREC2012} 5th Workshop on the Representation and Processing of Sign Languages: Interactions between Corpus and Lexicon}, pages 109--116, Istanbul, Turkey. {European Language Resources Association (ELRA)}.

\bibitem[{Lualdi et~al.(2023)Lualdi, Spiecker, Wooten, and Clark}]{lualdi2023advancing}
Colin~P Lualdi, Barbara Spiecker, Alicia~K Wooten, and Kaitlyn Clark. 2023.
\newblock Advancing scientific discourse in american sign language.
\newblock \emph{Nature Reviews Materials}, 8(10):645--650.

\bibitem[{Mathias et~al.(2022)Mathias, Sarah, Cihan, Zifan, Alessia, Amit, Annette, Richard, and Ryan}]{muller_mathias_2022_6621480}
Müller Mathias, Ebling Sarah, Camgöz~Necati Cihan, Jiang Zifan, Battisti Alessia, Moryossef Amit, Rios Annette, Bowden Richard, and Wong Ryan. 2022.
\newblock \href {https://doi.org/10.5281/zenodo.6621480} {Wmt-slt focusnews: Training data for the wmt shared task on sign language translation}.

\bibitem[{Mauk and Tyrone(2012)}]{mauk2012location}
Claude~E Mauk and Martha~E Tyrone. 2012.
\newblock Location in asl: Insights from phonetic variation.
\newblock \emph{Sign Language \& Linguistics}, 15(1):128--146.

\bibitem[{Mauk(2003)}]{mauk2003undershoot}
Claude~Edward Mauk. 2003.
\newblock \emph{Undershoot in two modalities: Evidence from fast speech and fast signing}.
\newblock The University of Texas at Austin.

\bibitem[{Napoli et~al.(2011)Napoli, Sanders, and Wright}]{napoli2011some}
Donna~Jo Napoli, Nathan Sanders, and Rebecca Wright. 2011.
\newblock Some aspects of articulatory ease in american sign language.
\newblock \emph{Handout from Stony Brook University, May}, 6:2011.

\bibitem[{Napoli et~al.(2014)Napoli, Sanders, and Wright}]{napoli2014linguistic}
Donna~Jo Napoli, Nathan Sanders, and Rebecca Wright. 2014.
\newblock On the linguistic effects of articulatory ease, with a focus on sign languages.
\newblock \emph{Language}, 90(2):424--456.

\bibitem[{Padden and Perlmutter(1987)}]{padden1987american}
Carol~A Padden and David~M Perlmutter. 1987.
\newblock American sign language and the architecture of phonological theory.
\newblock \emph{Natural language \& linguistic theory}, 5(3):335--375.

\bibitem[{Piantadosi et~al.(2011)Piantadosi, Tily, and Gibson}]{piantadosi2011word}
Steven~T Piantadosi, Harry Tily, and Edward Gibson. 2011.
\newblock Word lengths are optimized for efficient communication.
\newblock \emph{Proceedings of the National Academy of Sciences}, 108(9):3526--3529.

\bibitem[{Radford et~al.(2021)Radford, Kim, Hallacy, Ramesh, Goh, Agarwal, Sastry, Askell, Mishkin, Clark, Krueger, and Sutskever}]{clip}
Alec Radford, Jong~Wook Kim, Chris Hallacy, Aditya Ramesh, Gabriel Goh, Sandhini Agarwal, Girish Sastry, Amanda Askell, Pamela Mishkin, Jack Clark, Gretchen Krueger, and Ilya Sutskever. 2021.
\newblock \href {https://api.semanticscholar.org/CorpusID:231591445} {Learning transferable visual models from natural language supervision}.
\newblock In \emph{International Conference on Machine Learning}.

\bibitem[{Shaw and Anthony(2016)}]{shaw}
Alex Shaw and Lisa Anthony. 2016.
\newblock \href {https://doi.org/10.1145/2993148.2993179} {Analyzing the articulation features of children's touchscreen gestures}.
\newblock In \emph{Proceedings of the 18th ACM International Conference on Multimodal Interaction}, ICMI '16, page 333–340, New York, NY, USA. Association for Computing Machinery.

\bibitem[{Sigurd et~al.(2004)Sigurd, Eeg-Olofsson, and Van~Weijer}]{sigurd2004word}
Bengt Sigurd, Mats Eeg-Olofsson, and Joost Van~Weijer. 2004.
\newblock Word length, sentence length and frequency--zipf revisited.
\newblock \emph{Studia linguistica}, 58(1):37--52.

\bibitem[{Spearman(1961)}]{spearman1961proof}
Charles Spearman. 1961.
\newblock The proof and measurement of association between two things.

\bibitem[{Stamp et~al.(2022)Stamp, Khatib, and Hel-Or}]{stamp2022capturing}
Rose Stamp, Lilyana Khatib, and Hagit Hel-Or. 2022.
\newblock Capturing distalization.
\newblock In \emph{Proceedings of the LREC2022 10th Workshop on the Representation and Processing of Sign Languages: Multilingual Sign Language Resources}, pages 187--191.

\bibitem[{Starner et~al.(2023)Starner, Forbes, So, Martin, Sridhar, Deshpande, Sepah, Shahryar, Bhardwaj, Kwok et~al.}]{starner2023popsign}
Thad Starner, Sean Forbes, Matthew So, David Martin, Rohit Sridhar, Gururaj Deshpande, Sam Sepah, Sahir Shahryar, Khushi Bhardwaj, Tyler Kwok, et~al. 2023.
\newblock Popsign asl v1. 0: An isolated american sign language dataset collected via smartphones.
\newblock \emph{Advances in Neural Information Processing Systems}, 36:184--196.

\bibitem[{Stokoe~William(1960)}]{stokoe1960sign}
C~Stokoe~William. 1960.
\newblock Sign language structure: An outline of the visual communication systems of the american deaf (studies in linguistics occasional papers 8) buffalo.
\newblock \emph{NY: University of Buffalo}.

\bibitem[{Supalla(1978)}]{supalla1978many}
Ted Supalla. 1978.
\newblock How many seats in a chair?
\newblock \emph{Understanding language through sign language research}.

\bibitem[{Tanzer et~al.(2024)Tanzer, Shengelia, Harrenstien, and Uthus}]{tanzer-etal-2024-reconsidering}
Garrett Tanzer, Maximus Shengelia, Ken Harrenstien, and David Uthus. 2024.
\newblock \href {https://doi.org/10.18653/v1/2024.emnlp-main.360} {Reconsidering sentence-level sign language translation}.
\newblock In \emph{Proceedings of the 2024 Conference on Empirical Methods in Natural Language Processing}, pages 6262--6287, Miami, Florida, USA. Association for Computational Linguistics.

\bibitem[{Trujillo et~al.(2020)Trujillo, Simanova, Bekkering, and {\"O}zy{\"u}rek}]{trujillo2020communicative}
James~P Trujillo, Irina Simanova, Harold Bekkering, and Asli {\"O}zy{\"u}rek. 2020.
\newblock The communicative advantage: How kinematic signaling supports semantic comprehension.
\newblock \emph{Psychological research}, 84(7):1897--1911.

\bibitem[{Tyrone and Mauk(2010)}]{tyrone2010sign}
Martha~E Tyrone and Claude~E Mauk. 2010.
\newblock Sign lowering and phonetic reduction in american sign language.
\newblock \emph{Journal of Phonetics}, 38(2):317--328.

\bibitem[{Tyrone and Mauk(2012)}]{tyrone2012phonetic}
Martha~E Tyrone and Claude~E Mauk. 2012.
\newblock Phonetic reduction and variation in american sign language: A quantitative study of sign lowering.
\newblock \emph{Laboratory phonology}, 3(2):425.

\bibitem[{{Unity Technologies}(2023)}]{unity}
{Unity Technologies}. 2023.
\newblock \href {https://unity.com/} {Unity}.
\newblock Game development platform.

\bibitem[{Yin et~al.(2024)Yin, Regier, and Klein}]{yin-etal-2024-american}
Kayo Yin, Terry Regier, and Dan Klein. 2024.
\newblock \href {https://doi.org/10.18653/v1/2024.acl-long.839} {{A}merican {S}ign {L}anguage handshapes reflect pressures for communicative efficiency}.
\newblock In \emph{Proceedings of the 62nd Annual Meeting of the Association for Computational Linguistics (Volume 1: Long Papers)}, pages 15715--15724, Bangkok, Thailand. Association for Computational Linguistics.

\bibitem[{Zhang et~al.(2020)Zhang, Bazarevsky, Vakunov, Tkachenka, Sung, Chang, and Grundmann}]{zhang2020mediapipe}
Fan Zhang, Valentin Bazarevsky, Andrey Vakunov, Andrei Tkachenka, George Sung, Chuo-Ling Chang, and Matthias Grundmann. 2020.
\newblock Mediapipe hands: On-device real-time hand tracking.
\newblock \emph{arXiv preprint arXiv:2006.10214}.

\bibitem[{Zhou et~al.(2021)Zhou, Zhou, Qi, Pu, and Li}]{zhou2021improving}
Hao Zhou, Wengang Zhou, Weizhen Qi, Junfu Pu, and Houqiang Li. 2021.
\newblock Improving sign language translation with monolingual data by sign back-translation.
\newblock In \emph{Proceedings of the IEEE/CVF Conference on Computer Vision and Pattern Recognition}, pages 1316--1325.

\bibitem[{Zipf(2016)}]{zipf2016human}
George~Kingsley Zipf. 2016.
\newblock \emph{Human behavior and the principle of least effort: An introduction to human ecology}.
\newblock Ravenio books.

\end{thebibliography}

\section{Language Resource References}
\label{lr:ref}
\bibliographystylelanguageresource{lrec2026-natbib}
\bibliographylanguageresource{languageresource}

\appendix
\section{Full List of STEM Signs}
\label{app:full_STEM_list}
This appendix summarizes the STEM vocabulary used in both the isolated vocabulary condition described in \S~3.1. Table~\ref{tab:stem_variants} reports the unique STEM terms included in the dataset and the number of distinct signed variants observed for each term.

\begin{table}[h]
\centering
\small
\begin{tabular}{l c}
\toprule
\textbf{STEM Term} & \textbf{\# Distinct Signed Variants} \\
\midrule
System & 3 \\
Energy & 2 \\
Species & 3 \\
Theory & 2 \\
Cell & 3 \\
Plant & 2 \\
Carbon & 2 \\
Gas & 3 \\
Iron & 2 \\
Mass & 2 \\
Chemical & 2 \\
Element & 2 \\
Oxygen & 5 \\
Organism & 2 \\
Properties & 3 \\
Photosynthesis & 4 \\
Chloroplast & 1 \\
Chlorophyll & 1 \\
Carbon Dioxide & 2 \\
Glucose & 2 \\
Autotrophs & 1 \\
Chemical Equation & 4 \\
Oxidation & 2 \\
Reduction & 2 \\
Cell Respiration & 2 \\
Mitosis & 5 \\
Meiosis & 1 \\
DNA & 1 \\
DNA Replication & 1 \\
DNA Transcription & 1 \\
DNA Translation & 2 \\
\bottomrule
\end{tabular}
\caption{Unique STEM terms and the number of distinct signed variants observed in the dataset. Variants include fingerspelled forms and annotated phonological alternatives.}
\label{tab:stem_variants}
\end{table}

\section{Keypoint Mapping}
To ensure comparability between the 3D motion capture data and 2D MediaPipe outputs, we manually defined a correspondence between the motion capture joint names and MediaPipe’s landmark indices. Table~\ref{tab:joint_mapping} lists the specific MediaPipe keypoints corresponding to each motion capture joint for both the right and left sides of the body.
\label{app:joint_mapping}
\begin{table}[h!]
\centering
\footnotesize
\resizebox{\linewidth}{!}{%
\begin{tabular}{lll}
\toprule
 & \textbf{Joint Name} & \textbf{Mediapipe Keypoint} \\
\midrule
\multirow{23}{*}{Right} 
 & RightArm & pose\_12 \\
 & RightForeArm & pose\_14 \\
 & RightHand & pose\_16 \\
 & RightHandMiddle1 & right\_hand\_9 \\
 & RightHandMiddle2 & right\_hand\_10 \\
 & RightHandMiddle3 & right\_hand\_11 \\
 & RightHandMiddle4 & right\_hand\_12 \\
 & RightHandRing & right\_hand\_13 \\
 & RightHandRing1 & right\_hand\_14 \\
 & RightHandRing2 & right\_hand\_15 \\
 & RightHandRing4 & right\_hand\_16 \\
 & RightHandPinky & right\_hand\_17 \\
 & RightHandPinky1 & right\_hand\_18 \\
 & RightHandPinky2 & right\_hand\_19 \\
 & RightHandPinky4 & right\_hand\_20 \\
 & RightHandIndex & right\_hand\_5 \\
 & RightHandIndex1 & right\_hand\_6 \\
 & RightHandIndex2 & right\_hand\_7 \\
 & RightHandIndex4 & right\_hand\_8 \\
 & RightHandThumb1 & right\_hand\_1 \\
 & RightHandThumb2 & right\_hand\_2 \\
 & RightHandThumb3 & right\_hand\_3 \\
 & RightHandThumb4 & right\_hand\_4 \\
\midrule
\multirow{23}{*}{Left} 
 & LeftArm & pose\_11 \\
 & LeftForeArm & pose\_13 \\
 & LeftHand & pose\_15 \\
 & LeftHandMiddle1 & left\_hand\_9 \\
 & LeftHandMiddle2 & left\_hand\_10 \\
 & LeftHandMiddle3 & left\_hand\_11 \\
 & LeftHandMiddle4 & left\_hand\_12 \\
 & LeftHandRing & left\_hand\_13 \\
 & LeftHandRing1 & left\_hand\_14 \\
 & LeftHandRing2 & left\_hand\_15 \\
 & LeftHandRing4 & left\_hand\_16 \\
 & LeftHandPinky & left\_hand\_17 \\
 & LeftHandPinky1 & left\_hand\_18 \\
 & LeftHandPinky2 & left\_hand\_19 \\
 & LeftHandPinky4 & left\_hand\_20 \\
 & LeftHandIndex & left\_hand\_5 \\
 & LeftHandIndex1 & left\_hand\_6 \\
 & LeftHandIndex2 & left\_hand\_7 \\
 & LeftHandIndex4 & left\_hand\_8 \\
 & LeftHandThumb1 & left\_hand\_1 \\
 & LeftHandThumb2 & left\_hand\_2 \\
 & LeftHandThumb3 & left\_hand\_3 \\
 & LeftHandThumb4 & left\_hand\_4 \\
\bottomrule
\end{tabular}%
}
\caption{Mapping between motion capture joints and Mediapipe keypoints.}
\label{tab:joint_mapping}
\end{table}

\section{Avatar Examples}
\begin{figure}[t]
    \centering
    \begin{minipage}{0.48\linewidth}
        \centering
        \includegraphics[width=\linewidth]{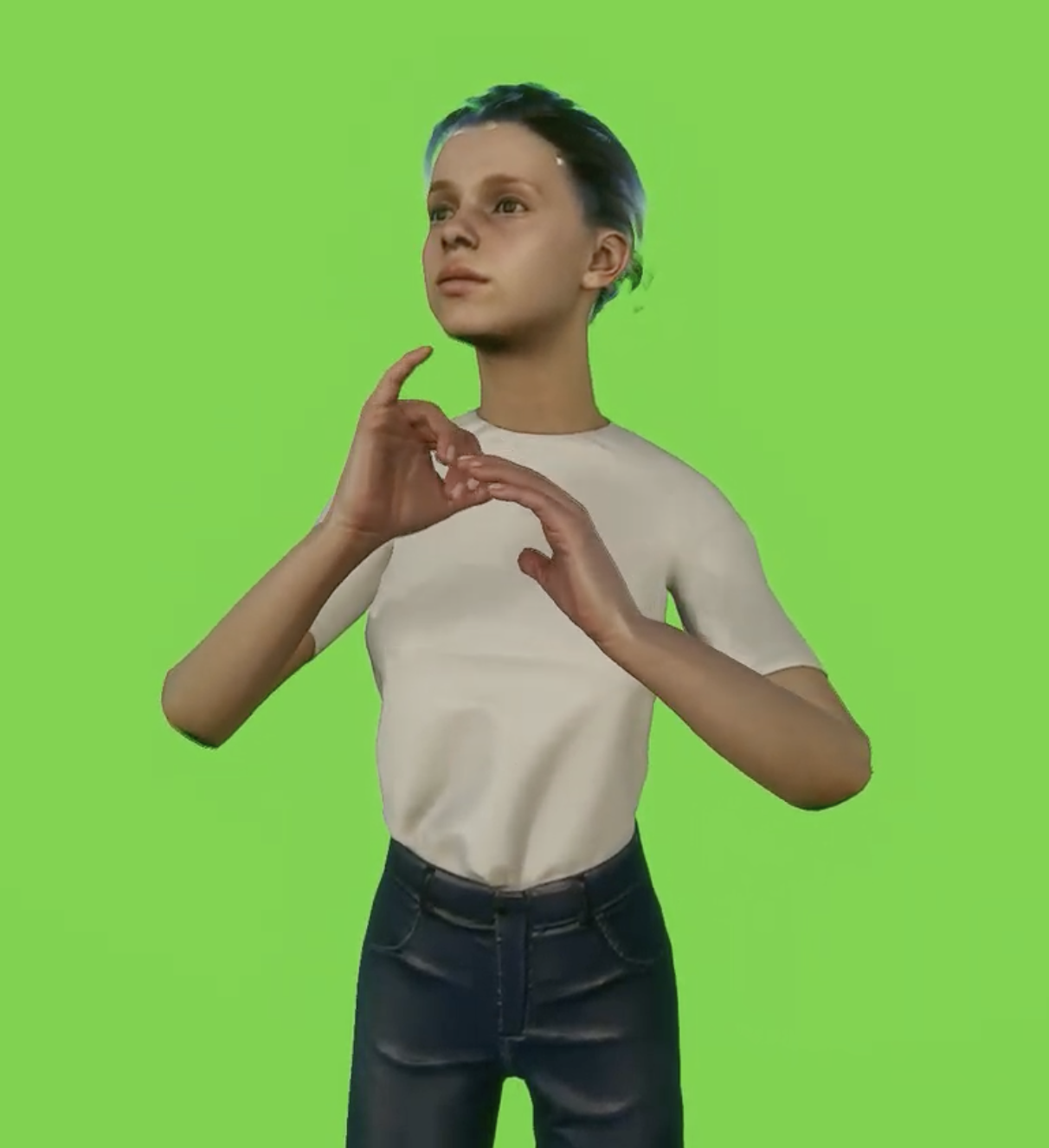}
    \end{minipage}
    \hfill
    \begin{minipage}{0.48\linewidth}
        \centering
        \includegraphics[width=\linewidth]{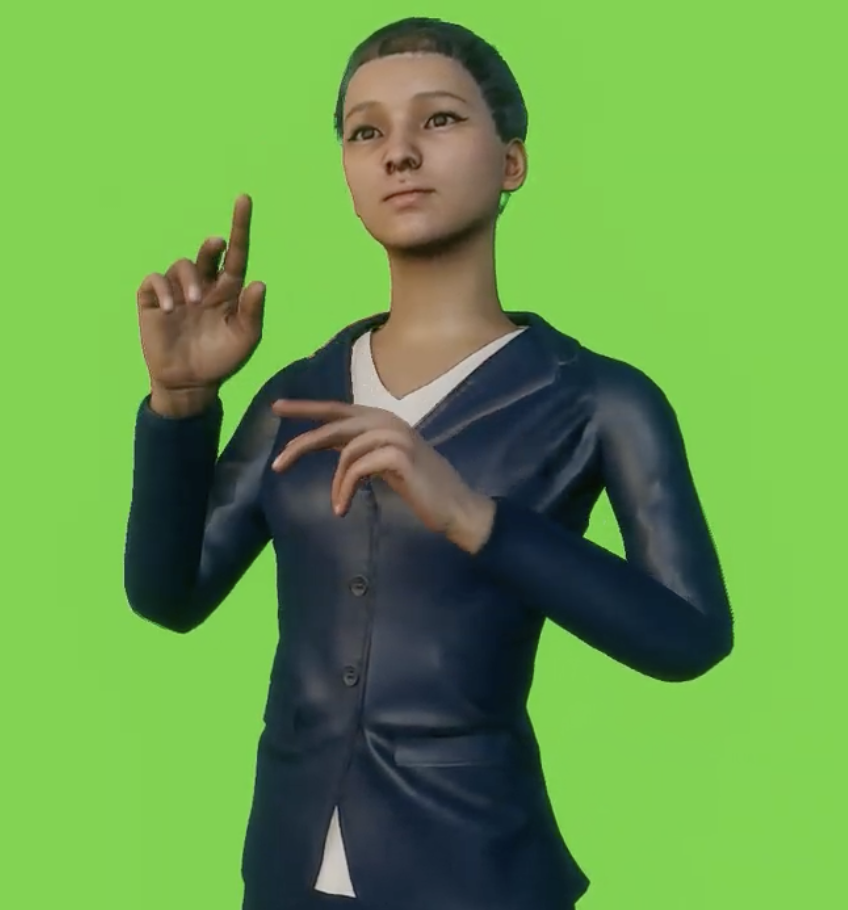}
    \end{minipage}
    \caption{Example frames from avatar rendered motion capture recordings.}
    \label{fig:avatars}
\end{figure}

Figure~\ref{fig:avatars} shows example frames from the avatar rendered versions of the
motion capture recordings used in our experiments. Using Unity, we rendered 3D humanoid avatars driven by the recorded 3D joint trajectories. Avatar rendered videos were used exclusively for evaluating pretrained sign language models (\S~3.4), in order to assess whether reducing out-of-distribution visual noise improves model performance.

\label{app:avatar_examples}

\section{Relative Change Analysis - Student}
\label{app:student_results}

This section reports the relative change analysis for the student signer, conducted using the same method described in \S~\ref{ssec:repeated_mention}. For each joint, we computed Spearman rank correlations between mention index and the percentage change in spatial extent, path length, and velocity across repeated mentions of the same sign. Positive values indicate articulatory reduction (i.e., smaller movements or increased efficiency) with repetition. 

As shown in Table~\ref{tab:relative_change_student}, compared to the instructor, the student produced fewer repeated signs, which limited the statistical power of within-sign analyses. None of the correlations reached significance. This suggests no consistent pattern of reduction across repetitions for the student signer in the dialogue condition.

\renewcommand{\arraystretch}{1.2}
\begin{table}[ht]
\centering
\resizebox{\linewidth}{!}{
\begin{tabular}{llll}
\toprule
\textbf{Joint} & \textbf{Spatial} & \textbf{Path} & \textbf{Velocity} \\
\midrule
Fingers (L)  & $+0.231$ & $+0.137$ & $+0.171$ \\
\rowcolor{gray!20} Fingers (R) & $+0.334$ & $+0.352$ & $-0.297$ \\
Hand (L)      & $+0.309$ & $+0.309$ & $-0.160$ \\
\rowcolor{gray!20} Hand (R)    & $+0.302$ & $+0.322$ & $-0.290$ \\
Forearm (L)   & $+0.137$ & $-0.005$ & $+0.265$ \\
\rowcolor{gray!20} Forearm (R) & $-0.071$ & $-0.154$ & $-0.070$ \\
Arm (L)       & $+0.188$ & $+0.147$ & $+0.114$ \\
\rowcolor{gray!20} Arm (R)     & $-0.030$ & $-0.233$ & $+0.114$ \\
\bottomrule
\end{tabular}
}
\caption{Relative Change Analysis (Spearman correlation). 
Stars denote significance (* $p<.05$, ** $p<.01$, *** $p<.001$). 
None of the correlations reached significance. }
\label{tab:relative_change_student}
\end{table}


\end{document}